\documentclass[conference]{IEEEtran}
\IEEEoverridecommandlockouts
\usepackage{cite}
\usepackage{amsmath,amssymb,amsfonts}
\usepackage{algorithmic}
\usepackage{graphicx}
\usepackage{todonotes}
\usepackage{booktabs}
\usepackage{array}
\usepackage{caption} 
\usepackage{subfigure}
\usepackage{bm}
\usepackage{multirow}
\usepackage{tabularx,colortbl}
\usepackage{hyperref}
\usepackage{balance}

\usepackage{textcomp}
\def\BibTeX{{\rm B\kern-.05em{\sc i\kern-.025em b}\kern-.08em
    T\kern-.1667em\lower.7ex\hbox{E}\kern-.125emX}}

\begin{document}

\title{Time Series Kernel Similarities for Predicting Paroxysmal Atrial Fibrillation from ECGs}

\author{
\IEEEauthorblockN{Filippo Maria Bianchi\IEEEauthorrefmark{2}\IEEEauthorrefmark{1}, Lorenzo Livi\IEEEauthorrefmark{4}, Alberto Ferrante\IEEEauthorrefmark{2}, Jelena Milosevic\IEEEauthorrefmark{3}, Miroslaw Malek\IEEEauthorrefmark{2}}
\IEEEauthorblockA{\IEEEauthorrefmark{2}\textit{ALaRI, Faculty of Informatics, Universit\`a della Svizzera italiana}, \textit{Lugano, Switzerland}\\
\IEEEauthorrefmark{1}\textit{Machine Learning Group, UiT the Arctic University of Norway}, \textit{Troms\o{}, Norway}\\
\IEEEauthorrefmark{3}\textit{Institute of Telecommunications, TU Wien, Vienna, Austria}\\
\IEEEauthorrefmark{4}\textit{Dept. of Computer Science, University of Exeter, Exeter, UK}\\
\textit{Email: \IEEEauthorrefmark{1}filippo.m.bianchi@uit.no, \IEEEauthorrefmark{4}l.livi@exeter.ac.uk}, \\\IEEEauthorrefmark{2}\{alberto.ferrante,miroslaw.malek\}@usi.ch, \IEEEauthorrefmark{3}jelena.milosevic@tuwien.ac.at}
}

\maketitle

\begin{abstract}
We tackle the problem of classifying Electrocardiography (ECG) signals with the aim of predicting the onset of Paroxysmal Atrial Fibrillation (PAF).
Atrial fibrillation is the most common type of arrhythmia, but in many cases PAF episodes are asymptomatic. 
Therefore, in order to help diagnosing PAF, it is important to design procedures for detecting and, more importantly, predicting PAF episodes.
We propose a method for predicting PAF events whose first step consists of a feature extraction procedure that represents each ECG as a multi-variate time series.
Successively, we design a classification framework based on kernel similarities for multi-variate time series, capable of handling missing data.
We consider different approaches to perform classification in the original space of the multi-variate time series and in an embedding space, defined by the kernel similarity measure.
We achieve a classification accuracy comparable with state of the art methods, with the additional advantage
of detecting the PAF onset up to 15 minutes in advance.
\end{abstract}


\section{Introduction}

Wearable devices are nowadays able to capture bio-signals, such as electrocardiograms (ECGs), for an extended period of time. 
Data recorded by these devices is of paramount clinical importance in the assessment of numerous heart-related conditions. Among them, the prediction of Paroxysmal Atrial Fibrillation (PAF) episodes and the risk stratification of PAF-prone patients is of high importance  \cite{alcaraz2010review,hickey2004non,milosevic2014risk}.
Atrial fibrillation is the most common type of arrhythmia. 
Its symptoms include fatigue or decreased exercise tolerance, palpitations, dyspnea on exertion, and generalized weakness, even though in many cases PAF episodes are asymptomatic \cite{PAF-symptoms}. 
The risk of PAF progressing to permanent atrial fibrillation increases with time.

The problem of detecting and predicting PAF events from ECG recordings received attention from researchers from different fields \cite{di2013characteristics}.
Many approaches have been proposed to tackle such problems, including the use of entropy descriptors \cite{lo2015outlier,cervigon2010entropy}, P-wave characterization \cite{martinez2015alteration,dotsinsky2007atrial,alcaraz2013morphological}, the number of premature atrial complexes from R-R intervals \cite{1275571} (i.e., the interval between R peaks, shown in Fig. \ref{fig:ecg_waves}), heart variability rate \cite{schlenker2016recurrence}, the average number of f-waves in a TQ interval \cite{du2014novel}, and hybrid complex networks / time series analysis techniques \cite{li2012bridging,li2011detection}.

In this work, we propose a new framework based on time series analysis methods to predict the onset of PAF.
The overall procedure consists of different steps.
First, from each ECG we extract numerical descriptors (features) for characterizing the original signals.
Then, we perform feature selection to extract a reduced subset of features according to their mutual dependencies.
The values assumed over time by the selected features are represented as a multivariate time series (MTS); that might contain missing values.
Finally, in order to predict the PAF onset, we train a classifier based on special kernel similarity measures designed for MTS with missing data~\cite{Baydogan2016, mikalsen2017time}.
The classifier is trained either in the MTS input space, using directly the relationships among data defined by the kernel similarity, or in an embedding space where each MTS is represented by its similarity values with respect to the training data.

To evaluate the effectiveness of the proposed framework, we take into account the data from the PhysioNet PAF prediction challenge \cite{paf,moody2001predicting,goldberger2000physiobank}.
Results show performance higher or on par with respect to the state-of-the-art results to correctly identify ECGs associated with PAF events. 
We also have the option to predict the PAF onset up to 15 minutes before the event, thus providing the ability to alert the patient of the possible occurrence of a PAF episode.

Our work provides the following key contributions:
\begin{itemize}
    \item As methodological contribution, we propose a novel method for predicting PAF events that is based on a classifier trained either on the input or in the embedding space induced by a kernel similarity measure. In such a space, samples are described only by their similarities with respect to the elements in the training set.
    \item Kernel similarities for MTS with missing data~\cite{Baydogan2016, mikalsen2017time} are a recent development in time series analysis and they have never been considered so far in the context of PAF prediction, which represents thus a new field of application.
    \item We analyze a new type of PAF prediction problem, by studying how much time ahead the PAF events can be accurately detected by considering only the earliest measurements in the ECGs.
    \item The proposed framework is characterized by a high degree of parallelization. This provides the opportunity of significantly speeding up the computation if the procedure is implemented on highly scalable systems, such as embedded devices with parallel computing support.
\end{itemize}

The remaining part of the paper is organized as follows: Section \ref{sec:dataset} describes the dataset of the PAF prediction challenge that we have used to develop our method; Section \ref{sec:methodology} describes the details of our framework, and Section \ref{sec:experiments} presents the results that we have obtained.

\section{Dataset description}
\label{sec:dataset}

We consider the ECG data from the Physionet PAF prediction challenge \cite{paf,moody2001predicting,goldberger2000physiobank}.
Half of the records present in the database are acquired from patients prone to PAF (either immediately before fibrillation or after some time a fibrillation occurred), the other half from healthy subjects. 
No ECG of patients with an ongoing PAFs is considered.
The database provides a training set, consisting of 50 records of PAF patients and 50 records of healthy subjects and a test set containing 50 records. 
Excerpts are 30 minutes long, concurrently acquired from two sensors.

The PAF prediction challenge~\cite{paf} aims at predicting the onset of PAF events before their occurrence. 
Therefore, the task we tackle here is not a simple detection problem.
The dataset under analysis consists of 106 recordings organized as follows.
For the test set of the first PAF event (PAF detection), there are 22 recordings classified as ``control'' and 28 as ``PAF-prone''.
For generating the test set of the PAF prediction challenge (the one considered in this paper), only the 28 recordings of PAF-prone patients are considered.
These 28 PAF-prone recordings are further divided into two classes: the ``close-to-onset'' and ``far-from-onset'' class. The former, represents records preceding a PAF event; the latter, instead, represents the records that do not precede a PAF event.
Therefore, for generating the test set of the PAF prediction challenge, we end up with 56 recordings equally distributed between the two different classes.
The training set is constructed by considering again pairs of recordings, this time taken from the 25 PAF-prone subjects belonging to the records used for training -- for a total of 50 records to be used for training.

\section{Methodology}
\label{sec:methodology}

\begin{figure}[th!]
	\centering
	\includegraphics[width=\columnwidth]{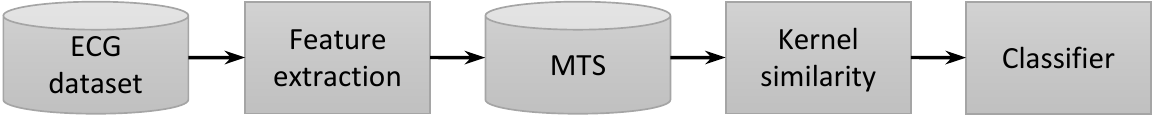}
	\caption{Schematic depiction of the proposed classification procedure. The original dataset of ECG signals is processed by a feature extraction block, to represent them as MTS. Specific kernel measures are then used to evaluate the similarities between the MTS, which are stored in a kernel matrix. Such kernel matrix is finally exploited to train a classifier for predicting the PAF onset for each original ECG signal.}
	\label{fig:PAF_schema}
\end{figure}

We propose a classification framework, which implements the procedure summarized in Fig.~\ref{fig:PAF_schema}.
The procedure consists of three main steps: i) feature extraction, ii) computation of the kernel similarity and iii) training of the classifier.
The details of each phase are given in the remainder of this section.

\subsection{Feature extraction}
\label{sec:feat_extract}

The objective of this step is to extract from each ECG a set of descriptors, which are useful for the classification.
Each selected descriptor is represented by a variable in an MTS that, in turn, becomes the new representation of the original ECG.
Therefore, the ECG dataset will be represented by $N$ MTS $\mathbf{X}_1, \dots, \mathbf{X}_N$ with $\mathbf{X} \in \mathbb{R}^{V \times T}$, where $V$ is the number of variables, $T_m$ is the number of time steps of the shortest MTS, and $T_M$ the number of time steps of the longest one.
For the PAF dataset under analysis, we have $N=106$, $T_m=992$, and $T_M=3364$.
The number of variables $V$ is determined by the feature extraction procedure, which consists of two main steps: \textit{feature generation} and \textit{feature selection}.

\subsubsection{Feature generation}

\begin{figure}[th!]
	\centering
	\subfigure[]{
	\includegraphics[width=.5\columnwidth]{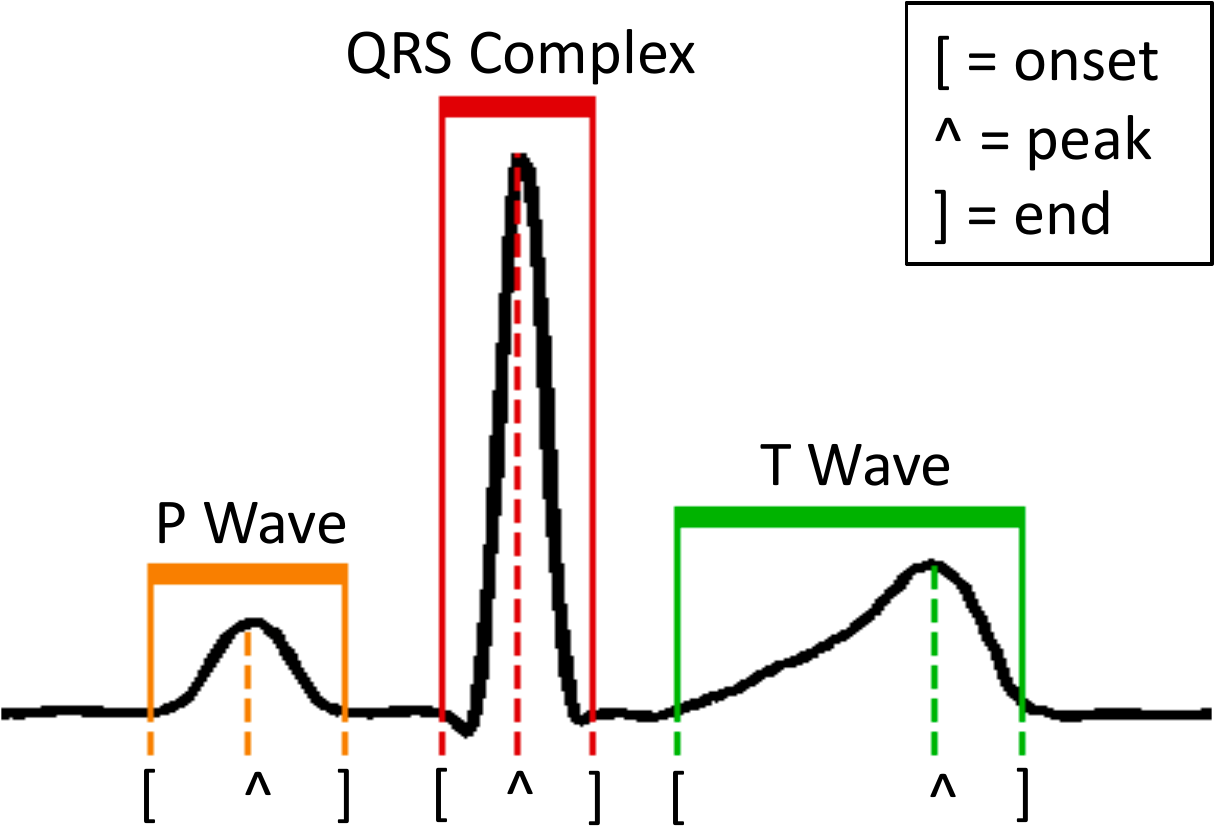}}
	~
	\subfigure[]{
	\includegraphics[width=.42\columnwidth]{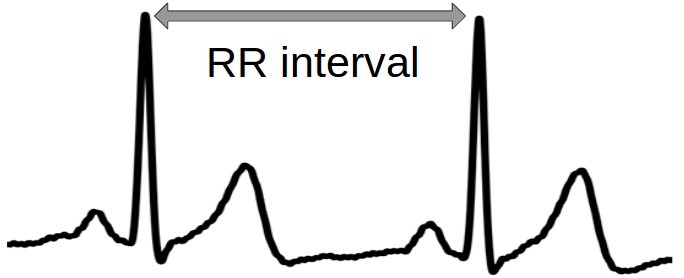}}
	\caption{In panel (a) the fiducial points of an ECG heartbeat \cite{milosevic2014risk}. In panel (b), an example of an R-R interval, corresponding to the time span between the R peaks of two consecutive QRS complex.}
	\label{fig:ecg_waves}
\end{figure}

The ECG signals are initially processed for identifying the interval between two successive heartbeats (the R-R interval shown in Fig.~\ref{fig:ecg_waves}). 
The variation in the time interval between heartbeats provides important clinical descriptors, which are used to characterize heart failure and other circulatory diseases~\cite{malik2002relation}.

To compute the R-R intervals, we first pre-process the ECG signal using the morphological filtering technique described in \cite{EPFL-CONF-181118}. 
Morphological filtering allows to retain useful information from acquired signals, while effectively eliminating noise originating from multiple sources such as low-frequency baseline wandering (caused by respiration and perspiration) and higher-frequency components (caused by muscular activity). 
A further step to enhance the quality of the acquired data is to combine signals from different inputs (leads) before the feature extraction step. 
In this study, we employ a Root Mean Square (RMS) combination of the two signals provided in the PAF-prediction database. Delineation is performed both on the two individual signals provided by the PAF prediction database and on their root-mean-square combination. 

After pre-processing is completed, an algorithm based on the digital wavelet transform~\cite{EPFL-CONF-181118,milosevic2014risk} is used to retrieve R-R interval as well as the the fiducial points of each heartbeat: the start, peak, and end of each characteristic wave (see Fig. \ref{fig:ecg_waves}).

There may be cases in which fiducial points cannot be extracted due to the low quality of the ECG signal. 
Low quality may be due to different reasons, e. g. imprecise/faulty measurements registered by the leads.
In those cases, there will be a missing value in the descriptors, in correspondence of the time steps relative to portion of ECG measured with low quality.

While other techniques with better performance are available for extracting the fiducial points, the ones used in this work are suitable for being used in resources-constrained devices, such as the wearable ones. Thus, they satisfy the conditions in which our methodology is designed to operate.

The output of this procedure produces, for each original ECG signal, $95$ time series of features related to morphological wave characteristics (for details on such features, see \cite{milosevic2014risk}).

\subsubsection{Feature selection}
\begin{figure}[th!]
	\centering
	\subfigure[]{
	\includegraphics[keepaspectratio,width=0.5\columnwidth]{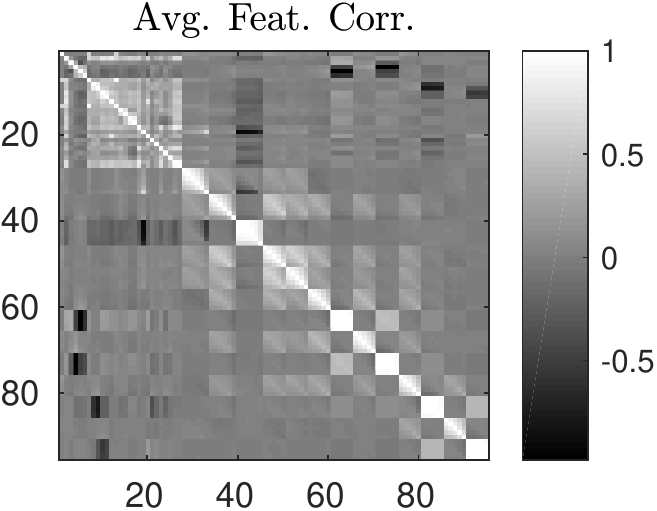}
	\label{fig:corrM_plain}
	}
	~
	\subfigure[]{
	\includegraphics[keepaspectratio,width=0.4\columnwidth]{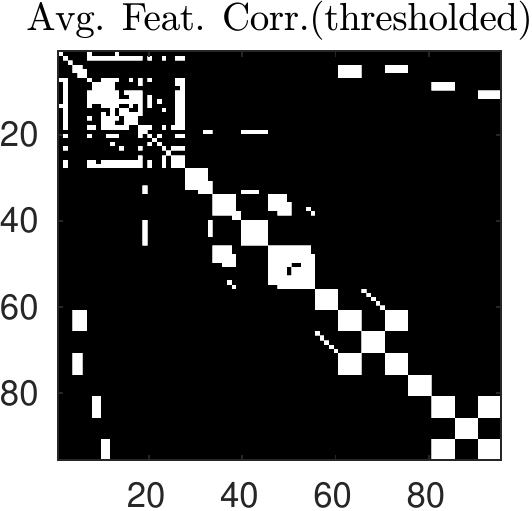}
	\label{fig:corrM_thresh}
	}
	\caption{In (a), the average correlation matrix of the ECG features for the MTS in the whole dataset. In (b), the same matrix where the values have been binarized using $\theta_c = 0.4$.}
	\label{fig:corrM}
\end{figure}

After the feature generation procedure, each original ECG is represented by an MTS with $V=95$ variates. 
Initially, we considered all the available features.
However, preliminary results showed that training a classifier using all variates yielded worse results.
Indeed, most of the original features are highly correlated.
Furthermore, by including all features, the computational time to evaluate the kernel similarities and the training of the classifier increase considerably.
Therefore, we considered different subsets of features in our experiments by performing feature selection on the original 95 variates.
For guiding the selection of the features, we evaluated their linear Pearson correlation.
In Fig.~\ref{fig:corrM}, we show the sample correlation matrix $\mathbf{C}$ between the 95 variables, which is computed as follows:
\begin{enumerate}
    \item $c_n^{i,j} = \mathrm{corr}(\mathbf{x}_n^i, \mathbf{x}_n^j)$;
    \item $\bar{c}^{i,j} = \frac{1}{N} \sum_n c_n^{i,j}$;
    \item $\mathbf{C} = \begin{cases}
    \mathbf{C}^{i,j} = 0 \; \text{if} \; |\bar{c}^{i,j}| < \theta_c,\\
    \mathbf{C}^{i,j} = 1 \; \text{otherwise}.
    \end{cases}$
\end{enumerate}
where $\mathbf{x}_n^i$ is the time series relative to the $i$-th variable in the $n$-th MTS $\mathbf{X}_n$ and $\theta_c$ is a hyperparameter that thresholds the maximum degree of correlation between the variables. 

In particular, we avoided to select at the same time two attributes if they were highly correlated (depicted as white pixels in the correlation matrix in Fig. \ref{fig:corrM}).
As a suitable threshold for the average correlation matrix $\mathbf{C}$, we used $\theta_c = 0.4$. This choice allows to get rid of highly correlated variables and, at the same time, significantly reduce the memory requirements of the procedure.
After this feature selection procedure, we reduced the number of variables to $V=22$. 
Those, describe features that are mostly related to R-R intervals and are summarized in Tab.~\ref{tab:feats}.
%

\bgroup
\def\arraystretch{0.95} 
\setlength\tabcolsep{.2em} 
\begin{table}[!ht]
\small
\centering
\caption{Variables identified by the feature selection procedure; see Fig. \ref{fig:ecg_waves} for reference.}
\label{tab:feats}
\begin{tabular}{|p{0.38\columnwidth}|p{0.55\columnwidth}|}
\hline
\rowcolor{gray!30} \textbf{Name} & \textbf{Description} \\
\hline
R\_Position & Position of the R peak \\
\hline
IdentifiedStructures & Number of structures in the beat (P, R, and Q waves) \\
\hline
R\_Max & Peak value of the R wave \\        
\hline
R\_Area & Area under the R wave \\    
\hline
R\_Width & Width of the R wave \\ 
\hline
R\_ini & Time distance from the beginning of the R peak until the maximum value \\         
\hline
P\_Position & Position of the P wave\\       
\hline
P\_Max & Peak value of the P wave \\         
\hline
P\_Area & Area under the P wave \\       
\hline
P\_Width & Width of the P wave \\     
\hline
P\_ini & Time distance from the beginning of the P peak until the maximum value \\
\hline
PR\_Interval & Interval between the P and the R waves\\  
\hline
T\_Position & Position of the T peak \\
\hline
T\_Max & Peak value of the T wave \\            
\hline
T\_Area &  Area under the T wave \\       
\hline
T\_Width &  Width of the T wave \\      
\hline
T\_ini & Time distance from the beginning of the T peak until the maximum value \\
\hline
RT\_Interval & Interval between the R and T wave \\         
\hline
RR\_Interval & Interval between two R peaks \\         
\hline
PR\_Segment & Time distance between P and R waves \\
\hline
RR\_Interval\_50HB\_Mean & Interval between two R peaks, mean over the last 50 samples \\
\hline
RR\_Interval\_5HB\_Mean & Interval between two R peaks, mean over the last 5 samples \\
\hline
\end{tabular}
\end{table}
\egroup

\subsection{Kernel similarities for multivariate time series}
\label{sec:ts_kernel}

Dynamic Time Warping \cite{berndt1994using} is the most widely used approach to assess the similarity between time series, but cannot treat MTS in its original formulation.
More complicated variants have been proposed to deal with multiple variables~\cite{shokoohi2017generalizing}, but they are characterized by difficult hyperparameter tuning and high computational complexity.
Furthermore, there is not a straightforward way to account for missing values in the MTS when computing the similarities.

In this work, we adopt two different kernel similarities for MTS, the \textit{Learned Pattern Similarity} (LPS)~\cite{Baydogan2016} and the \textit{Time series Cluster Kernel} (TCK)~\cite{mikalsen2017time}, whose details are provided in the following two subsections.
Those methods benefit from high parallelization capabilities and learn a model that, once the training is over, can quickly process new unseen data.
Both procedures for computing LPS and TCK require all the MTS to have equal length $T$.
Therefore, we followed the approach proposed in~\cite{Wang2016237} and, by means of linear interpolation, we transformed all MTS so that they have the same number of time steps $T$.
The value of $T$ is determined by 
$
T = \left \lceil \frac{T_{M}}{\left \lceil \frac{T_{M}}{25} \right \rceil} \right \rceil,
$
where $T_M$ is the length of the longest MTS in the dataset and $ \lceil \: \rceil$ is the ceiling operator. 
Before computing TCK we also standardized each MTS, i.e., from each variable $v$ in the dataset we subtract its mean value $\bar{v}$ and divide by its standard deviation $\sigma_v$; both $\bar{v}$ and $\sigma_{v}$ are computed on the training set.
On the other hand, since decision trees are scale invariant, we did not apply this transformation for LPS (in accordance with~\cite{Baydogan2016}).

The training procedure returns a kernel matrix $\mathbf{K}_\text{tr} \in \mathbb{R}^{N_\text{tr} \times N_\text{tr}}$, whose components are the similarities among the elements of the training set.
Once LPS and TCK are fitted, we process the test set and generate an additional matrix $\mathbf{K}_\text{te} \in \mathbb{R}^{N_\text{te} \times N_\text{tr}}$, whose components are the similarities of the elements in the training set with respect to those in the test set.
$N_\text{te}$ and $N_\text{ts}$ represent the size of test and training set respectively, with $N = N_\text{te} + N_\text{ts}$.

TCK and LPS account for the missing patterns in the data to compute their similarities, rather than relying on imputation methods that may introduce strong biases. 
Imputation techniques replace the missing values with predefined or computed values, such as 0s, the empirical mean, or the last seen values.
The choice is often arbitrary and, in presence of large amounts of missing values, the resulting data may end up being significantly different as well as the result of the analysis.

\subsubsection{Learned Pattern Similarity}
\label{sec:lps}

The learned pattern similarity (LPS) algorithm~\cite{Baydogan2016} is based on the identification of segments-occurrence within time series.
It generalizes naturally to MTS by means of regression trees where a bag-of-words type compressed representation is created, which in turn is used to compute the similarity score.
LPS is considered the state-of-the-art for MTS, inherits the decision tree approach to handle missing data, and is based on an ensemble strategy.
Specifically, one extracts from MTS all possible segments of length $L(L < T)$ starting from each time index $t = 1,2, \dots, T-L+1$ and fit them to $nT$ randomly initialized regression trees.
Since LPS naturally deals with missing data, in general it is not necessary to replace the missing entries.

LPS depends on two hyperparameters, which are the maximum number of regression trees considered, $nT$, and the length of the segments, denoted as $L$.

\subsubsection{Time Series Cluster Kernel}
\label{sec:tck}

The \emph{Time series Cluster Kernel} (TCK) \cite{mikalsen2017time} computes an unsupervised kernel similarity for MTS with missing data.
The method is based on an ensemble learning approach wherein the robustness to hyperparameters is ensured by joining the clustering results of many Gaussian mixture models (GMM) to form the final kernel. Hence, no critical hyperparameters have to be tuned by the user.

In order to deal with missing data, the GMMs are extended using informative prior distributions \cite{Marlin:2012:UPD:2110363.2110408}.
To generate partitions with different resolutions that capture both local and global structures in the input data, the TCK similarity matrix is built by fitting GMMs to the set of time series for a number of mixtures, each one with a different number of components.
To provide diversity in the ensemble, each partition is evaluated on a random subset of attributes and segments, using random initializations and randomly chosen hyperparameters. 
This also helps to provide robustness for what concerns hyperparameters selection.
TCK is then built by summing (for each partition) the inner products between pairs of posterior distributions corresponding to different time series.

The hyperparameters that needs to be specified in TCK are the number of different random initializations $Q$ and the maximum number of Gaussian mixtures $C$.
It is sufficient to set those hyperparameters to reasonable high values to obtain good performance in many tasks~\cite{mikalsen2016learning, 2018arXiv180307879O}.

\subsection{Classification}
\label{sec:classification}

Once the kernel matrices are generated, we train a classifier based on the similarities yielded by $\mathbf{K}_\text{tr}$ and $\mathbf{K}_\text{te}$.
While in principle we can use any classifier operating on real-valued vectors, we consider two different ones, which are the $k$-Nearest Neighbors classifier ($k$NN) and the Support Vector Machine classifier (SVM).
To perform classification we follow two different approaches: \textit{classification in the input space} and \textit{classification in the similarity-induced embedding space}.

\subsubsection{Classification in input space}
In the first approach, classification is performed directly in the MTS input space, considering as similarity between the samples the values contained in the kernel matrix $\mathbf{K}_\text{te}$.
In $k$NN, for each element $\mathbf{X}_i$ in the test set we select the relative row $\mathbf{k}_i \in \mathbf{K}_\text{te}$ and then we select the indices relative to the $k$ highest values in $\mathbf{k}_i$, which identify the $k$ MTS of the training set that are most similar to $\mathbf{X}_i$.
The estimated $\mathbf{X}_i$ label $\hat{y}_i$ is the most frequent one among those $k$ MTS.

On the other hand, in SVM is possible to train the classifier using $\mathbf{K}_\text{tr}$ as precomputed kernel, which defines the kernel space where the optimal separating hyperplane is computed.
In this case, the label $\hat{y}_i$ is assigned according to the region (delimited by such hyperplane) that contains $\mathbf{X}_i$.

\subsubsection{Classification in embedding similarity space}
In this second approach, rather than assuming as input for the classification the original MTS, whose similarity is described by the kernel matrix, we train the classifier on the rows of $\mathbf{K}_\text{tr}$ and $\mathbf{K}_\text{te}$.
In particular, each row $\mathbf{k}_i \in \mathbb{R}^{N_\text{tr}}$ is considered as a vectorial embedding representation of the original MTS $\mathbf{X}_i$ in the similarity space.

Therefore, in this approach to train the classifier it is necessary to compute an additional vector similarity among the rows of both $\mathbf{K}_\text{tr}$ and $\mathbf{K}_\text{te}$.
In $k$NN, we consider Euclidean, cosine, Cityblock and Pearson correlation similarities to identify, for each vector in $\mathbf{K}_\text{te}$, the $k$ most similar vectors in $\mathbf{K}_\text{tr}$.
Analogously, in SVM we compute a kernel similarity among the rows of $\mathbf{K}_\text{tr}$ and $\mathbf{K}_\text{te}$ using radial basis functions (rbf), whose analytic expression reads
\begin{equation}
\label{eq:rbf}
d_{ij} = \mathrm{exp} \left\{ \gamma \| \mathbf{k}_i - \mathbf{k}_j \|^2 \right\},
\end{equation}
where $\gamma$ defines the bandwidth of the kernel.
By applying Eq.~\ref{eq:rbf} to all the rows $i \in \mathbf{K}_\text{tr}$ and $j \in \mathbf{K}_\text{te}$, the embedding vectors are mapped into a new kernel space, where the optimal separating hyperplane is computed.

Compared to directly performing the classification in the input space, this approach requires an additional computation to evaluate the vector similarities among the representations (rows/columns of the kernel matrix) of the original MTS.
The increment in the training time scales with the dimensionality of the vectors, which are equal to $N_\text{tr}$, i.e. the size of the training set.
The procedure is more computational demanding in the case a SVM classifier is adopted, since the rbf kernels must be evaluated for each pair of vectors.

On the other hand, the proposed approach based on similarity-induced embedding provides a stronger form of regularization, which usually improves the generalization capability of the classifier. 
Additionally, the similarity embedding can model better the relationships among classes, thanks to its robustness to instance level noise~\cite{weinberger2009large}.
Finally, the similarity embedding can highly synthesize the information when the MTS are characterized by a large number of variables and time steps, i.e. when $T \times V \gg N_\text{tr}$.
However, in those cases where the classifier is required to capture a higher degree of variance among the data, the embedding approach might not be beneficial.

\section{Results}
\label{sec:experiments}
In this section, we describe the experiments that we have performed and we report the results that we have obtained. 
The whole classification framework, including TCK and LPS kernels, are implemented in MATLAB\footnote{TCK: \url{https://github.com/kmi010/Time-series-cluster-kernel-TCK-}, LPS: \url{http://www.mustafabaydogan.com/learned-pattern-similarity-lps.html}}.
In each experiment, for LPS we set the number of regression trees to $nT = 200$ and the length of the segments to $L = 10$.
Instead, for TCK we set the maximum number of Gaussian mixtures to $C = 40$ and the number of random initializations to $Q = 30$.

\begin{figure}[th!]
	\centering
	\subfigure[LPS]{
	\includegraphics[height=4cm, width=4cm]{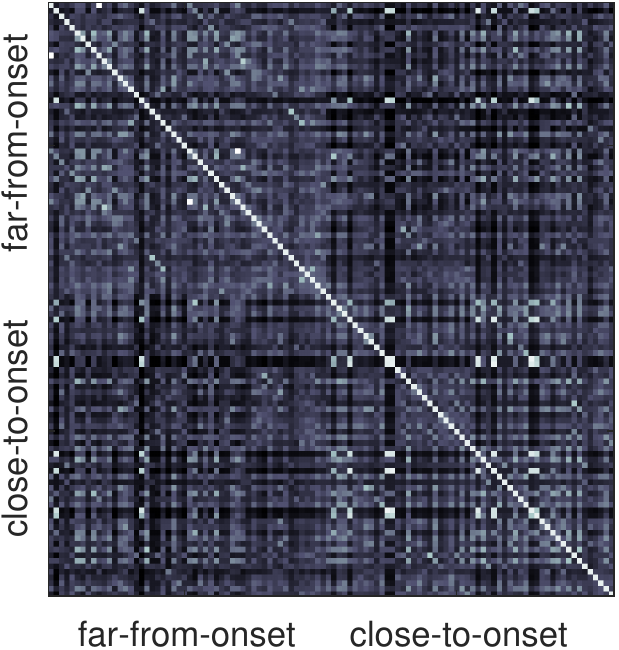}
	\label{fig:Klps}
	}
	~
	\subfigure[TCK]{
	\includegraphics[height=4cm, width=4cm]{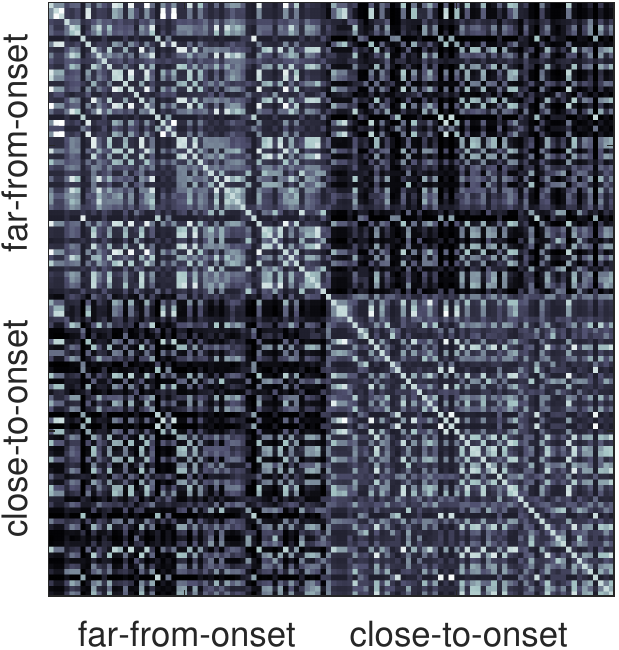}
	\label{fig:Ktck}
	}
	\caption{Kernel matrix obtained with the LPS and TCK algorithms.}
	\label{fig:KernelMatrix}
\end{figure}

The two kernel matrices yielded by TCK and LPS are depicted in Fig.~\ref{fig:KernelMatrix}.
The higher the values at the position $(i,j)$ in such matrices, the higher the similarity between the corresponding MTS $\mathbf{X}_i$ and  $\mathbf{X}_j$.
In the depiction, brighter colors correspond to higher similarities.
It is possible to observe a block structure in the matrices, which denotes that the intra-class similarities are higher than the inter-class similarities.
The blocks representing the MTS relative to \textit{far-from-onset} and \textit{close-to-onset} patients are visible more clearly in the kernel matrix yielded by TCK, which suggests its superior capability to discriminate between the two classes.

In the following, we present two different experiments.
First, in Sec.~\ref{sec:paf2class} we report the classification results on the original PAF dataset relative to the second challenge (see Sec.~\ref{sec:dataset} for details), obtained by means of the proposed framework. 
Then, in Sec.~\ref{sec:windows} we analyze how such results change as we vary the number of time steps taken into account for the signals; accordingly, this results in evaluating how the performance deteriorates when we omit data closer to the actual PAF onset. 
This last experiment is especially interesting as it shows how our methodology can be used to predict a PAF episode before its occurrence.

\subsection{Classification results}
\label{sec:paf2class}

Here, we evaluate the results obtained with the proposed approach to classify the MTS into the \textit{far-from-onset} and \textit{close-to-onset} classes, which is the objective of the second PAF challenge (PAF2)~\cite{paf}.
For comparison, we also report the three best official results, in terms of accuracy, obtained on the PAF2 challenge~\cite{pafscores}.
As discussed in Sec.~\ref{sec:classification}, it is possible to use the TCK and LPS kernel matrices directly as the pre-computed kernel in a kernel-based classifier, such as SVM, or to exploit the pairwise similarity scores in a $k$NN classifier operating in input space.
As a second approach, we consider each row of the kernel matrix as an $N_\text{tr}$-dimensional embedding vector.
Compared to their original representations, the samples are now highly compressed since we move from $\mathbb{R}^{V \times T}$ (size of the original MTS) to $\mathbb{R}^{N_\text{tr}}$ (length of the embedding vector), with $V \times T = 319580$ and $N_\text{tr} = 50$ in our dataset. 
In the embedding-based approach, several similarity measures have been used for evaluating the relationships of the embedding vectors $\mathbf{k}_i$.  

To evaluate classification performance, we consider accuracy (ACC), recall on the ``close-to-onset'' class (REC), and finally the F1 score.
To select the optimal hyperparameters of the classifiers, we perform $k$-fold cross-validation.
For the $k$NN classifier, we consider a range of values for the number of neighbors $k$ in $[1,49]$ and, in case of classification in the embedding space, we consider as possible vector similarities Euclidean distance, cosine similarity, CityBlock distance and Pearson correlation.
In presence of ties (when $k$ assumes even values), we resolve them by deterministically assigning label of class ``far-from-to-onset'' to the samples.
For the SVM classifier, we search the margin of the hyperplane in $C \in [2^{-20}, 2^{10}]$ and the kernel bandwidth in $\gamma \in [2^{-5}, 2^5]$
The classification results obtained from the kernel similarities yielded by LPS and TCK are shown in Tab.~\ref{tab:PAF2_LPS_res}, along with the optimal hyperparameters found.
We refer to $k$NN-i, SVM-i and $k$NN-e, SVM-e as the $k$NN and the SVM applied in the input (-i) and embedding (-e) space, respectively.
\bgroup
\def\arraystretch{0.9} 
\setlength\tabcolsep{.3em} 
\begin{table}[!ht]
\footnotesize
\centering
\caption{Classification results in the input space ($k$NN-i and SVM-i) and in the embedding space ($k$NN-e and SVM-e), using the two time series similarities, LPS and TCK. 
We report classification accuracy (ACC), recall (REC) for the close-to-onset class and the F1 measure (F1) relative to the configuration achieving the best ACC value in the validation procedure.
We also report the optimal hyperparameters for such a configuration.
Best results for each method are shown in bold. 
The last entries in the table are the best official results reported so far for the challenge~\cite{pafscores}. Note that only ACC is reported as performance measure.}
\label{tab:PAF2_LPS_res}
\begin{tabular}{lllllllll}
\cmidrule[1.5pt]{1-9}
\textbf{Method} & \textbf{Similarity} & \textbf{ACC} & \textbf{REC} & \textbf{F1} & \textbf{$k$} & \textbf{Diss} & $\boldsymbol{C}$ & $\boldsymbol{\gamma}$ \\
\cmidrule[.5pt]{1-9}
\multirow{2}{*}{\textbf{$k$NN-i}} & LPS & \textbf{0.714} & \textbf{0.928} & \textbf{0.764} & 41 & -- & -- & -- \\
& TCK & 0.696 & 0.75 & 0.712 & 33 & -- & -- & -- \\
\cmidrule[.5pt]{1-9}
\multirow{2}{*}{\textbf{$k$NN-e}} & LPS & 0.66 & 0.678 & 0.666 & 49 & cosine & -- & -- \\
& TCK & 0.\textbf{678} & \textbf{0.683} & \textbf{0.675} & 29 & cityblock & -- & -- \\
\cmidrule[.5pt]{1-9}
\multirow{2}{*}{\textbf{SVM-i}} & LPS & 0.625 & \textbf{0.678} & 0.644 & -- & -- & 7.029 & -- \\
& TCK & \textbf{0.696} & 0.571 & \textbf{0.653} & -- & --  & 3.031 & -- \\
\cmidrule[.5pt]{1-9}
\multirow{2}{*}{\textbf{SVM-e}} & LPS &0.625 & 0.39 & 0.517 & -- & -- & 20.072 & 1.045 \\
& TCK & \textbf{0.632} & \textbf{0.643} & \textbf{0.631} & -- & -- & 1.749 & 0.075 \\
\cmidrule[0.5pt]{1-9}
\cite{schreier2001automatic} & -- & 0.71 & -- & --  & --  & --  & --  & --  \\
\cite{de2001automated} & -- & 0.68 & -- & --  & --  & --  & --  & --  \\
\cite{maier2001screening} & -- & 0.68 & -- & --  & --  & --  & --  & --  \\
\cmidrule[1.5pt]{1-9} 
\end{tabular}
\end{table}
\egroup

We observe that $k$NN and SVM when configured with TCK yield most of the time better results.
However, when $k$NN is configured with LPS and it operates in the input space, it provides the best results, which are comparable with the state-of-the-art.
Indeed, the highest official accuracy obtained for this problem is 0.71 and $k$NN-i+LPS obtains 0.714.

Interestingly, we notice that the validation procedure for the $k$NN always identifies as optimal high values of the number of neighbors ($k$).
A high value of $k$ improves the robustness to noise of the classifier. This is particularly important in real-world datasets, such as the one considered in this work, especially when the boundaries between different classes are not very distinct \cite{duda2001pattern}.
With respect to the remaining hyperparameters, we notice that the optimal configuration found with cross-validation varies from case to case.

\subsection{Results on windows of increasing length}
\label{sec:windows}

In this second experiment, we study how the classification outcomes change as we take into account an increasing number of time steps for the input MTS.
In particular, we expand an initially small time window located far from the end of recording by iteratively including measurements that are closer to end and, hence, closer to the PAF onset.
This allows us to evaluate how much time ahead the PAF events can be predicted accurately in our framework.

\begin{figure}[ht!]
	\centering
	\includegraphics[keepaspectratio,width=\columnwidth]{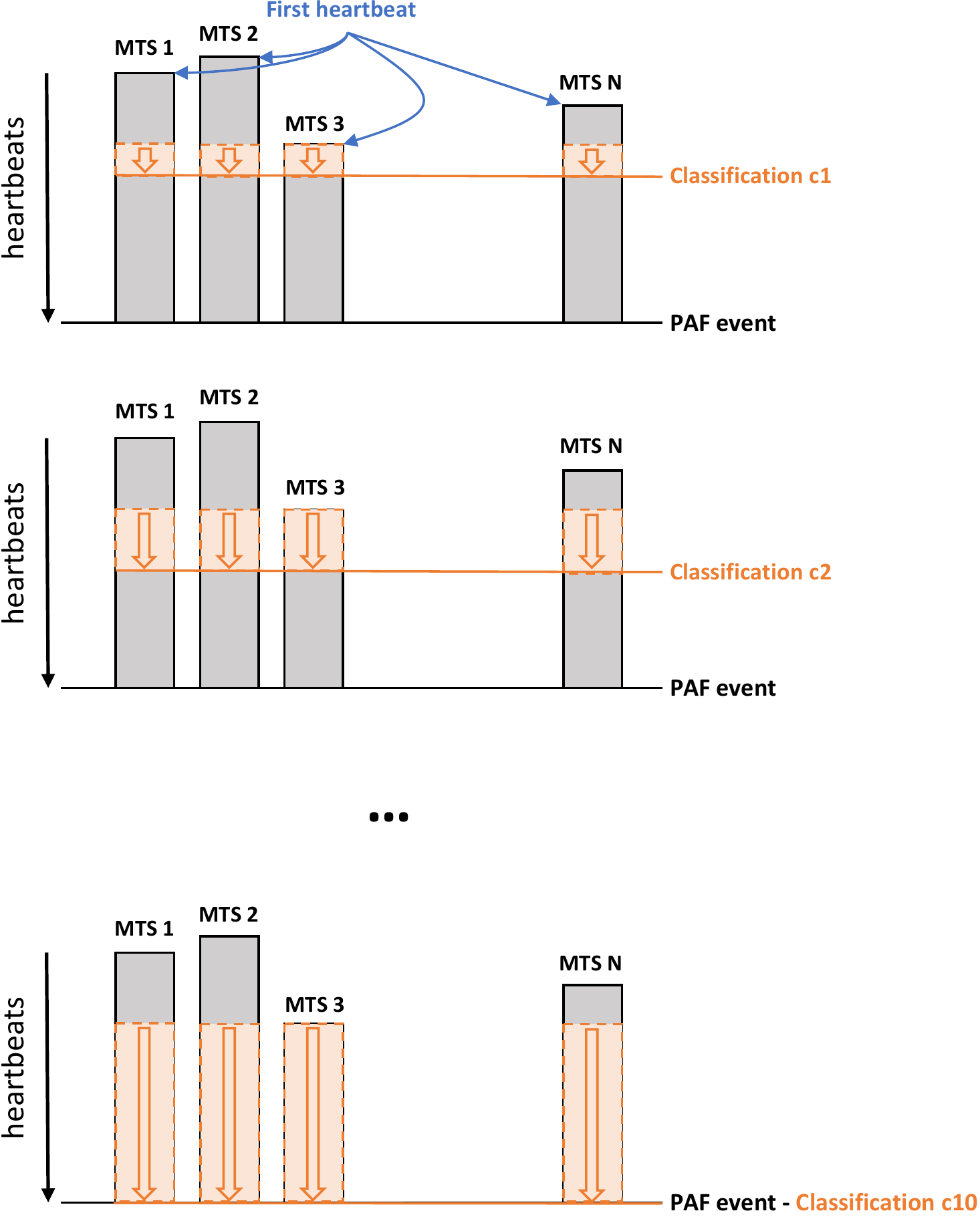}
	\caption{Classification on windows of increasing size. The first classification problem, c1, takes into account a small time-window of heartbeats, which is far from the actual PAF event. 
	The last classification problem, c10, takes into account the maximum number of time steps, by considering a window with a size equal to the shortest MTS in terms of heartbeats (MTS\textsubscript{3} in this case).}
	\label{fig:time_window}
\end{figure}

By applying a window of increasing size to the original data we obtain 10 different classification problems, c1, c2, \dots, c10.
Fig. \ref{fig:time_window} graphically illustrates the procedure.
The first problem c1 is created by considering the smallest window of heartbeats, positioned as far as possible from the PAF event (i.e., at the beginning of the recording).
Note that the first heartbeat considered is the first of the shortest time series (MTS\textsubscript{3} in the example shown in Fig.~\ref{fig:time_window}).
Therefore, the window of maximum width considered in the last classification problem c10 has length equal to the number of heartbeats in the shortest MTS of the PAF dataset (960 in this case).
The details of each one of the 10 classification problems are reported in Tab. \ref{tab:win_class}.
Note that, assuming that an adult has approximately 60-100 heartbeats per minute, the last column constitutes an approximation of the minutes between the end of the window and the occurrence of the PAF event.

\bgroup
\def\arraystretch{0.9} 
\setlength\tabcolsep{.3em} 
\begin{table}[!ht]
\footnotesize
\centering
\caption{Details of the classification problems generated by considering windows of increasing length. HB column denotes the number of heartbeats.}
\label{tab:win_class}
\begin{tabular}{cccc}
\cmidrule[1.5pt]{1-4}
\textbf{Problem} & \textbf{Win. size} & \textbf{HBs before PAF} & \textbf{Mins before PAF} \\
\cmidrule[.5pt]{1-4}
c1 & 96  & 864 & $\approx$ 9 - 15 \\
c2 & 192 & 768 & $\approx$ 8 - 13 \\
c3 & 288 & 672 & $\approx$ 7 - 11 \\
c4 & 384 & 576 & $\approx$ 6 - 10 \\
c5 & 480 & 480 & $\approx$ 5 - 8 \\
c6 & 576 & 384 & $\approx$ 4 - 7 \\
c7 & 672 & 288 & $\approx$ 3 - 5 \\
c8 & 768 & 192 & $\approx$ 2 - 3 \\
c9 & 864 & 96 & $\approx$ 1 - 2 \\
c10 & 960 & 0 & 0 \\
\cmidrule[1.5pt]{1-4}
\end{tabular}
\end{table}
\egroup

\begin{figure}[ht!]
	\centering
	\includegraphics[keepaspectratio,width=0.9\columnwidth]{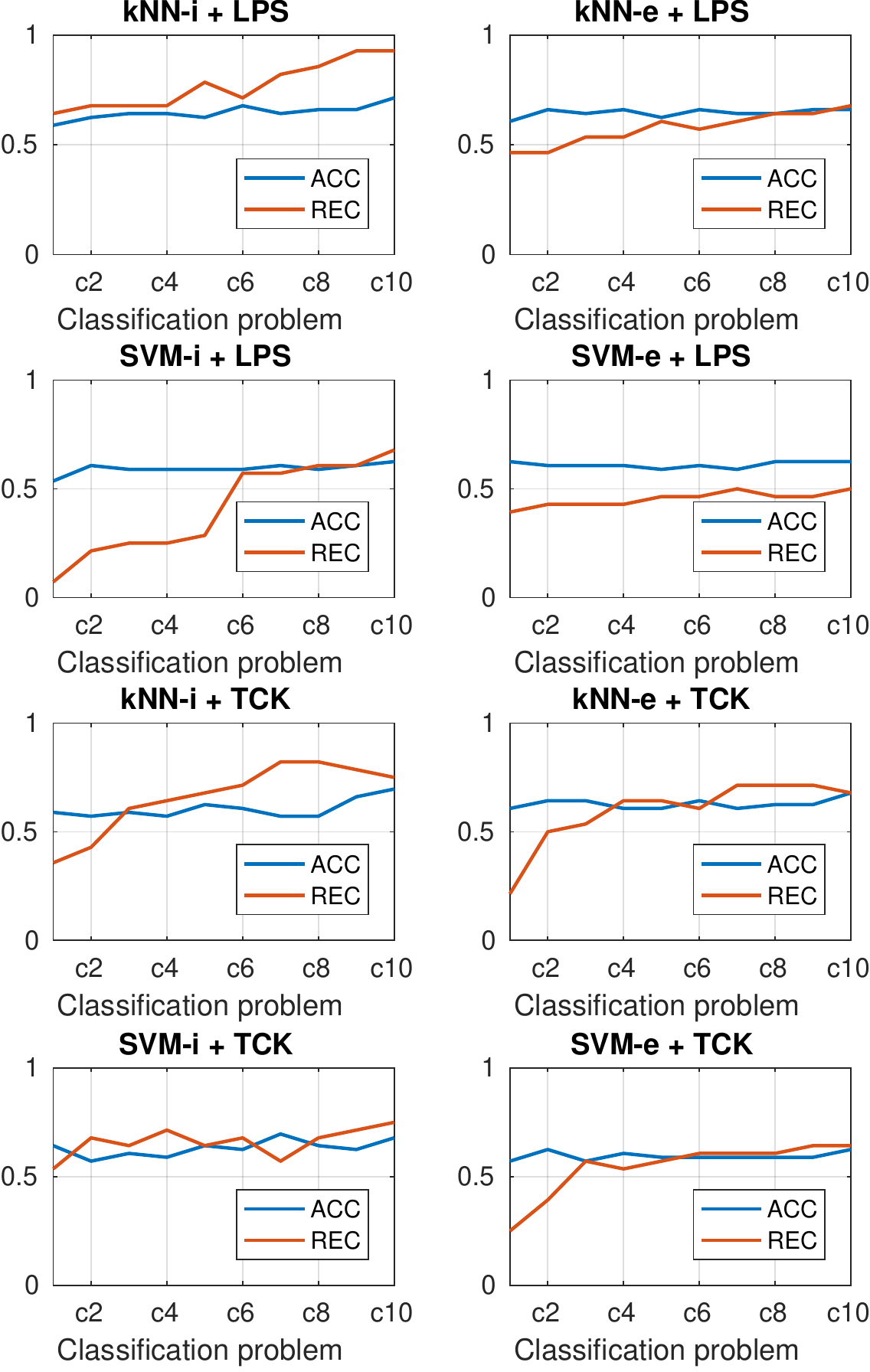}
	\caption{Classification accuracy (ACC) and recall (REC) for the close-to-onset class obtained by the different classifiers when considering 10 different windows of increasing size as reported in Table \ref{tab:win_class}.}
	\label{fig:time_window_res}
\end{figure}

The classification results for each problem by the different classification methods are reported in Fig.~\ref{fig:time_window_res}, in terms of accuracy (ACC) and recall (REC).
Each classifier uses the same hyperparameter configurations that are described in Tab.~\ref{tab:PAF2_LPS_res}.
We notice that the best performance for this task in terms of recall (REC) are obtained by the $k$NN classifier when is configured with LPS and operating in the input space.
On the other hand, when the $k$NN is operating in input space and it is configured with TCK, we obtain the best performance in terms of accuracy (ACC).
We also note that kNN-e configured with LPS achieves stable results for the different lengths of the time windows, since the performance of the classification does not deteriorate significantly when only a few data items before the event are considered.
This suggests that such a method can be the most suitable for an early detection of the PAF onset.
We note that, in this latter case, high accuracy and recall are obtained in the problem c5 where the considered time window terminates roughly 8-10 minutes before the actual PAF onset. 

Interestingly, in each experiment we observe that the recall is in general much lower when short time windows are considered.
On the other hand, the accuracy of the classification is very stable in each classification problem.
This demonstrates that the proposed method is capable of detecting the onset of PAF with a classification accuracy comparable with state of the art methods, with the additional advantage of detecting the PAF event several minutes before its occurrence.

\section{Conclusions}
\label{sec:conclusions}

In this paper, we proposed a methodology to classify ECG with the aim of predicting the onset Paroxysmal Atrial Fibrillation.
We represented the ECGs as multi-variate time series of measurements of different descriptors over time.
We have considered kernel functions for multi-variate time series, which allowed us to compute the pair-wise similarity between the ECGs.
Such similarities are then exploited for training a classifier, defined either in the original input space or in a similarity-induced embedding space.

Our classification results are higher or on pair with other state-of-the-art approaches. 
In addition, we show that with our proposed framework, it is possible to predict with a good precision the onset of PAF several minutes before the end of the available recordings.
This result is of significant practical importance, as it makes possible to conceive an early-warning system for the onset of PAF, which can alert a patient before the occurrence of the event.

The proposed methodology is particularly efficient in terms of computational resources.
In particular, the bottleneck, which is the computation of the kernel similarity matrix, can be implemented very efficiently by the TCK, which is an embarassingly parallelizable algorithm and its computational complexity scales down linearly with the number of available computing units.
Thanks to this property, the proposed framework is also suitable for implementations on highly parallel hardware devices and on multicore microprocessors, such as the ones that are being adopted in embedded systems.
Future includes porting onto embedded systems of the proposed software pipeline.


\section*{Acknowledgements}

This work was supported and funded by the Hasler Foundation under the Project ``HSD: A Personal Device for Automatic Evaluation of Health Status during Physical Training'' (Grant No. 15048). The paper reflects only the view of the authors.

The authors would like to thank Elisabetta De Giovanni, Dr. Amir Aminifar, and Prof. Alonso David Atienza from \'Ecole Polytechnique F\'ed\'erale de Lausanne for helping in the feature extraction process as well as for their precious feedback.

\balance
\bibliographystyle{IEEEtran}
\bibliography{Bibliography}

\end{document}